# Risk-based regulation for all: The need and a method for a wide adoption solution for data-driven inspection targeting.

**Celso Henrique Herédias Ribas**
*Digital Signal Processing Research Laboratory, Federal University of Santa Catarina, Santa Catarina, Brazil*
*Superintendence of Inspection, National Telecommunications Agency, Amazonas, Brazil*

**José Carlos Moreira Bermudez**
*Digital Signal Processing Research Laboratory, Federal University of Santa Catarina, Santa Catarina, Brazil*

**Abstract**
Access to data and data processing, including the use of machine learning techniques, has become significantly easier and cheaper in recent years. Nevertheless, solutions that can be widely adopted by regulators for market monitoring and inspection targeting in a data-driven way have not been frequently discussed by the scientific community. This article discusses the need and the difficulties for the development of such solutions, presents an effective method to address regulation planning, and illustrates its use to account for the most important and common subject for the majority of regulators: the consumer. This article hopes to contribute to increase the awareness of the regulatory community to the need for data processing methods that are objective, impartial, transparent, explainable, simple to implement and with low computational cost, aiming to the implementation of risk-based regulation in the world.

## 1. Introduction

All over the world, regulatory authorities face the same problem: It is impossible to control everything (Blanc 2012). Despite the billions in annual regulator budgets (European Court of Auditors 2020; US 2021), each regulator has limited resources, and most have a budget sufficient to inspect only a small part of the regulated entities (Johnson *et al.* 2020). As an alternative to overcome this limitation, risk-based regulation is widely discussed in international publications. Considering risk in formulating regulations and targeting regulator actions helps to improve all phases of regulatory policy (OECD 2021b). Rather than trying to inspect everything, risk-based regulation considers the probability of the non-compliance, as well as its consequences, in inspection resource allocation (OECD 2010), focusing efforts where better results can be achieved (Blanc 2012) and bringing efficiency, effectiveness and consistency to inspections (Borraz et al. 2020).

The literature on risk-based regulation and other widespread regulatory theories, such as responsive regulation (Ayres & Braithwaite 1992), discusses a lot about definitions of theoretical concepts associated with the proposed regulation, such as the definitions of non-compliance, transparency, efficiency and responsiveness, among others. Despite the progress made in recent years, however, most works fail to address the challenges faced by regulators on how to monitor the market to collect information, even before the suspicion of no compliance, to target inspections (Baldwin & Black 2008; Van Loo 2019). This makes the regulator's decision on how to target inspections even more difficult, often leading to heuristic approaches (Johnson *et al*. 2020). Regulator challenges include choosing which criteria or parameters should be used to define, classify, quantify and rank the risk. It is also necessary to define how to obtain the data, to store and manage them, as well as to determine how to use all the available information for planning and targeting inspections (Blanc 2012).

Monitoring the market is the first task of enforcement, preceding the tasks prescribed in Baldwin & Black (2008), without which no type of regulation will yield good results (Pecaric 2017). This task becomes more important when we realize that the monitoring-based decisions are often unobservable by the public and unreviewable by courts, and they can thus be more susceptible to improper influences (Van Loo 2019). There is consensus that random inspections and the attempt to inspect everything and everyone are not good options (Blanc 2012). Thus, it is necessary to face the challenges inherent to the phases of monitoring the market and targeting inspections so that the regulator resources are well allocated and the regulation achieves its objectives (Baldwin & Black 2008; Pecaric 2017). Often, it is monitoring, not enforcement, that makes regulated entities comply with rules (Campbell 2007).

New technologies have enabled improvements in regulatory activity, allowing the regulator to act more intelligently and proactively, monitoring remote, complex and widespread phenomena and objects (OECD 2021b). In a report published in September of 2021, the Organisation for Economic Co-operation and Development claims that regulators can use data analysis techniques to better target their inspections, and that easier and cheaper access to data and data processing, combined with inspector knowledge, will allow inspections in a more targeted way (OECD 2021a).

Digital processing of massive data (big data) through machine learning and artificial intelligence techniques has been increasingly applied to different problems (Lalmuanawma *et al*. 2020; Helm *et al*. 2020; Ullah *et al*. 2020), including analyses related to regulation (Johnson *et al*. 2020), government services and complaints (Chen *et al*. 2018). However, much has been discussed about ethics in the use of machine learning algorithms, mainly due to their potential to create and reproduce discrimination, and the difficulty for explaining these methods (Matus & Veale 2022, and references within). This theme has been the subject of some special issues in renowned publications in the area of policy, regulation and governance (Telecommunications Policy 2020; Regulation & Governance 2022). This highlights the importance of leveraging new technologies as tools to support good regulatory practices, instead of using them as substitutes for good practices (OECD 2021b).

To use data processing methods in regulation, it is necessary to ensure that the data are reliable and the processing methods are objective, impartial, transparent, explainable and simple to implement. Reliance on data and the objectivity of its processing is mandatory for risk-based regulation, as one must assess risk objectively and through data (OECD 2021b). Objectivity avoids discretionary decisions. Impartiality is necessary to avoid possible ethical deviations, while transparency and explainability are necessary properties to analyze and verify the trustworthiness of the method. Explainable methods, along with the simplicity of implementation, low computational cost and availability of reliable data, are needed for wide adoption by regulators and acceptance by inspectors, since changing inspector culture and improving their skills is a key for implementing risk-based regulation (OECD 2010). Despite the hype about machine learning techniques, methods with the aforementioned characteristics, aimed to be widely adopted by regulators for solving the problems of market monitoring and inspection targeting are still largely unavailable in the specialized literature. The entire regulatory community would benefit from the development of new methods that contribute to improve the best practices for implementing risk-based regulation.

Regulators are often subject to criticism about their capture by politicians or by the market, and about the waste of taxpayer money (Johnson *et al*. 2020). Proper inspection targeting contributes to an impartial and effective resource allocation, improving the image that society has of the regulator, since inspections are the basis for this judgment (Pecaric 2017). The effectiveness of regulators also increases the awareness of the population, and tends to increase its participation in the regulatory process (Ormosi 2012) by providing more and better data to the regulator, thus forming a virtuous cycle.

The aim of this article is to draw attention to the need for data processing methods that are objective, impartial, transparent, explainable and simple to implement, with low computational cost, in order to solve the problems of market monitoring and inspection targeting. It is our purpose to incentivize the proposition of new well debated methods for possible adoption by regulators around the world. This paper is organized as follows. Section 2 discusses the most important and common subject for the majority of regulators, namely, the consumer. Section 3 discusses how to contextualize consumer complaint data to analyze risk, and Section 4 proposes a method with the desired characteristics. Section 5 presents results of the application of the proposed method to the National Telecommunications Agency, the authority of the telecommunications market in Brazil, and Section 6 discusses the difficulties of using machine learning solutions and the advantages of the proposed method. Finally, Section 7 presents the conclusions of this work.

## 2. The most important and common subject for the majority of regulators

Each regulated sector or activity demands specific inspection techniques. However, some topics transcend sectors and activities and are common to the majority of regulators. The consumer policy is one of them. The United Nations determines that "Member States should work towards ensuring that consumer protection enforcement agencies have the necessary human and financial resources to promote effective compliance and to obtain or facilitate redress for consumers in appropriate cases" (UN 2016, p. 11), demonstrating the central importance of the topic. Although few regulators establish consumer standards, many of them enforce these standards (OECD 2021b), and promoting consumer satisfaction is a primary role for the majority of regulators.

The availability of consumer data is enormous in both national and transnational entities. Some examples of transnational entities are the International Consumer Protection and Enforcement Network, formed by consumer protection authorities from 70 countries, the European Consumer Centres Network, formed by consumer centers in all European Union countries and Norway, Iceland and United Kingdom, the Iberoamerican Forum of Consumer Protection Agencies, formed by the agencies of Latin America, Spain, and Portugal, the Coordinating Committee on Consumer Protection of the Association of Southeast Asian Nations, formed by 10 member states, among others.

The consumer data stored by regulators is varied, but most of the times includes consumer complaints data (OECD 2020). The very fact that these data are stored and available demonstrates the importance of the topic (Rottenburg *et al*. 2015). In addition to data of consumer complaints made to regulators, regulators also have access to data on consumer complaints made to regulated entities, which can be obtained through inspections (OECD 2020).

Several methods have been proposed in the literature to model and/or predict the quality of service for the consumer, considering services in general (Grönroos 1984; Parasuraman *et al*. 1985) as well as electronic, mobile and telecommunications services (Zeithaml etal. 2000; Fassnacht & Koese 2006;

Stiakakis & Georgiadis 2011; Kushwah & Bhargav 2014). The quality of service for the consumer depends on the disparity between the consumer perception about the service and the consumer expectation about the service (Parasuraman *et al*. 1985). Hence, consumer expectation about the service serves as a reference for their judgment (Fu *et al*. 2018).

Although consumer complaint data does not accurately quantify consumer dissatisfaction with the service, due to the various factors that influence whether or not a consumer chooses to complain, these complaint data reflect consumers expectations not met by the service (Europe Economics 2007) and, therefore, can be used as estimates of the quality of service for the consumer.

## 3. Contextualizing the data

The use of numerical indicators has intensified and gained importance in this century, changing the way government, organizations and people understand the world, due to their ability to simplify complex phenomena and allow these phenomena to be monitored, ranked and compared (Rottenburg *et al*. 2015). Academic articles and regulator publications use indicators and indices to quantify the performance of both the regulators themselves and of the markets and regulated entities (Mehrpouya & Samiolo 2019). The consumer complaint rate is one of the most publicized indices by regulators (ACMA 2022, Ofgem 2022, FTC 2022) and it is often used to target inspections (Europe Economics 2007).

However, one must be careful when interpreting consumer complaint data, especially for inspection targeting (Europe Economics 2007). The ranking of consumer complaint rates does not inform the origin of the problems that are occurring, but rather where their effects are being sensed. For example, if the complaint rate of a company in municipality A is high, this indicates that consumers in that municipality are not having their expectations met. However, if the complaint rates are high in all municipalities in the same region as municipality A, it is reasonable to intuit that the problem is regional, not local, and that the complaint rate in municipality A is high due to this regional problem. Therefore, it is not possible to state that the problem sensed at municipality A originates in the municipality A itself just by analyzing the complaint rate of that municipality. When segmenting the analysis of complaint rates by municipality, what is wanted is to highlight municipalities where local problems are occurring in order to target inspections to these municipalities.

Adequate models to process the data and transform it into risk quantification are essential (Rottenburg *et al*. 2015). In order to highlight the municipalities where local problems are occurring, it is necessary to contextualize the data so that the resulting metric indicates in which municipalities the complaint rates are higher than the rates expected to be observed, considering the context in which each municipality is inserted. The contextualization must quantify the percentage of influence that the complaint rates of the other municipalities exert on the formation of the expected value for the complaint rate of any given municipality. By carrying out this contextualization for all municipalities, an influence network is formed. With an influence network and the complaint rates, it is possible to define the quality of service for the consumer that the regulator expects to observe in each municipality, and then allocate inspections to those municipalities that present the greatest discrepancies between what is observed and what is expected by the regulator.

The influence network depends on the specific application that one wants to contextualize. The greater the knowledge of the application and the greater the amount of information available for contextualization, the more assertive the influence network will be. In the example of inspections directed at municipalities, one can model the connections and quantify the influences among municipalities in the formation of expected values of consumer complaint rates by considering the levels of interaction among the municipalities in the search for goods and services, as well as in command and control relations. Since the proposition of the central flow theory by Taylor *et al*. (2010), the number of researches on urban relations forming networks has increased and has modeled various phenomena of interaction between cities (Lüthi 2013; Papers in Regional Science 2016; Bettencourt 2021; Zhang & Tang 2021; Zhu 2021). Flows of people, resources, diseases and information have also been studied, especially using information and communications technologies (Steenbruggen 2015; Jia *et al.* 2020; Yamamoto *et al.* 2021), and can be used to contextualize the data in different applications. Influence networks can also be theoretically defined, as long as they model the expectation of the regulator from the data.

## 4. A solution for wide adoption

An area of digital data processing that has achieved remarkable theoretical and practical success in recent years is the Graph Signal Processing (GSP) (Marques *et al*. 2020). A graph is an abstract structure composed by a set of elements called vertices and edges. Vertices are the nodes of the graph, while edges connect pairs of vertices and describe inter-dependence relationships (Goldbarg & Goldbarg 2012) that can be used to contextualize the data on vertices. GSP deals with the development of tools and methods to process structured data (Puschel & Moura 2008a; 2008b; Shuman *et al*. 2013; cited on Gavili & Zhang 2017).

The signal domain is not always built out of equidistant time instants or of a regular grid in space. The data observation domain is irregular in numerous practical applications, and not always related to the time or space. Even when data is observed regularly in time and space, the consideration of new relations between the observation points may introduce new important information for the analysis and data processing. The data domain, in these cases, may be represented by a graph, and the data can be processed using graph signal processing algorithms (Stanković *et al*. 2019).

GSP has been developed following two major approaches (Gavili & Zhang 2017; Ortega *et al*. 2018; Bohannon *et al*. 2019). The first approach is based on algebraic signal processing, and mainly uses the adjacency matrix as Graph Shift Operator (Sandryhaila & Moura 2013; 2014a; 2014b), which serves as a model for the signal on graph (Ortega *et al*. 2018). The second approach is based on the definition of frequency spectra and expansion bases for a Graph Fourier Transform, and mainly uses the Laplacian matrix of graphs as the fundamental building block for its definitions and tools (Shuman *et al*. 2013). Both approaches define fundamental concepts of GSP (Gavili & Zhang 2017).

A weighted directed graph $\mathcal{G} = (\mathcal{V}, \mathcal{W})$ is a graph where $\mathcal{V}$ is the set of $N$ vertices of $\mathcal{G}$ and $\mathcal{W}$ is the set of edges that connect, uniquely in each orientation, two distinct vertices contained in $\mathcal{V}$. The adjacency matrix of $\mathcal{G}$ is defined as $\boldsymbol{W} \in \mathbb{R}^{N \times N}$ with elements $[\boldsymbol{W}]_{ij} = w_{ij}$, where $w_{ij} > 0$ is the weight of the edge connecting vertex $v_j \in \mathcal{V}$ to vertex $v_i \in \mathcal{V}$, going from $v_j$ to $v_i$ (Sardellitti *et al*. 2017). In this case, vertex $v_j$ is said to be the predecessor of vertex $v_i$, which in turn is said to be the successor of $v_j$. If there is no edge connecting $v_j$ to $v_i$ then $w_{ij} = 0$. A signal on an $N$-vertex graph is defined as $\boldsymbol{x} \in \mathbb{R}^N$ where $\boldsymbol{x} = [x_1, x_2, ..., x_N]^T$ is a vector whose element $x_i$ represents the information related to the vertex $v_i$.

The in-degree $g_i$ of a vertex $v_i$ in a graph is defined as the sum of the weights of the edges incident to that vertex (Sardellitti *et al*. 2017), i.e.

$$g_i = \sum_j w_{ij} \qquad (1)$$

The in-degree matrix of a graph is a diagonal matrix $G$ with main diagonal elements $[G]_i$ equal to the in-degrees of the corresponding vertices $v_i$, i.e.

$$G = \text{diag}(g_1, g_2, ..., g_N) \quad (2)$$

The Laplacian matrix of a graph is defined as

$$L = G - W \quad (3)$$

If the weights of all edges converging to each graph vertex are normalized so that its in-degree is equal to 1, the transformation of the graph signal by the graph Laplacian leads to the following expression for the $i$-th component of the resulting vector:

$$[Lx]_i = x_i - y_i \quad (4)$$

where $y_i$ is the estimator

$$y_i = \sum_j w_{ij} x_j \quad (5)$$

In vector form, the estimator vector is

$$y = Wx \quad (6)$$

For a deeper understanding of graph signal processing, we refer the reader to the articles by Sandryhaila and Moura (2013; 2014a; 2014b) and Shuman *et al.* (2013) and references therein.

For consumer complaint analysis, an adjacency matrix $W$ can be used to model the influence network between the vertices of the graph $\mathcal{G}$. In this way $w_{ij}$ quantifies the influence of $v_j$ on $v_i$. The graph signal $x$ can be a vector whose element $x_i \in \mathbb{R}^+$ represents the consumer complaint rate at municipality $v_i$. Considering the influences of all predecessor vertices of each graph vertex normalized so that they add up to 1, the transformation of the graph signal by the Laplacian graph will result, for each municipality, in the difference between the observed value of the consumer complaint rate and the expected value by the regulated for this rate.

However, we note that the difference between the consumer complaint rate and its predicted value does not reveal important information to determine the quality of service for the regulator. For example, if the consumer complaint rate is 2 complaints per million consumers and the predicted value for it is 1 complaint per million consumers, the difference between them is 1. If the consumer complaint rate were 101 complaints per million consumers and the predicted value for it was 100 complaints per million consumers, the difference between them would also be 1. Hence, the situations where the consumer complaint rate at a municipality is 100% or 1% worse than its predicted value result in the same information, demonstrating the need for a non-linear metric that yields an information which is more relevant for the application. We propose to use a metric that defines the discrepancy as the ratio between the consumer complaints rate and its predicted value. We conjecture that in several applications a metric that calculates the ratio $x_i / y_i$, where $y_i$ represents the predicted value for $x$, will be more suitable to quantify the discrepancy than a metric that calculates the difference $x_i - y_i$. We note that this relative metric is necessarily a nonlinear function of the graph signal at predecessor vertices. Such nonlinear metric cannot correspond to the transformation of the graph signal by the Laplacian. So, we define a discrepancy of graph vertex $v_i$ as

$$d_i = x_i / y_i \tag{7}$$

Defining the vertex discrepancy vector as $\boldsymbol{d} = [d_1, d_2, ..., d_N]^T$, with $d_i$ given by equation (7), $\boldsymbol{d}$ is expressed in vector form by

$$\boldsymbol{d} = \boldsymbol{x} \oslash \boldsymbol{y} \tag{8}$$

where $\oslash$ represents Hadamard division (element by element).

An interesting interpretation of equation (7) results if we combine it with equation (4) and write

$$d_i = ([\boldsymbol{Lx}] / [\boldsymbol{Wx}]_i) + 1 \tag{9}$$

As the value 1 in equation (9) is invariant with the graph signal, the vertex discrepancy $d_i$ can be interpreted as a normalized error metric, equivalent in information to the transformation of the graph signal by the Laplacian at a vertex, normalized by the weighted average of the signal amplitudes at its predecessor vertices. Unlike the Laplacian transformation, the vertex discrepancy is a nonlinear function of the graph signal.

The discrepancy can be taken as a measure of quality of service from the regulator's point of view, as it expresses the gap between the observed value of the consumer complaint rate and the regulator's expectation for this rate. Furthermore, the formulation of the solution through graphs allows the regulation community to benefit from a whole arsenal of tools developed for graph signal processing. More information about graph signal processing tools can be found, for instance, in the IEEE Signal Processing Magazine Special Issue on Graph Signal Processing, November 2020 (IEEE Signal Processing Magazine 2020).

## 5. Consumer complaint rates for inspection targeting

To illustrate the applicability of the proposed method, we analyzed the quality of the Personal Communications Service (PCS) by municipality and operator in Brazil, from January to December 2021, to indicate an order of priority for regulatory authority inspections for municipality level problems.

The National Telecommunications Agency (Anatel) is the independent regulatory authority for the Brazilian telecommunications market. On June 22, 2021, Anatel approved the Regulatory Enforcement Regulation (RER) (Anatel 2021b), whose rules are in line with risk-based and responsive regulation. According to Article 5 of the RER, the Regulatory Enforcement defines, among others, the following criteria: prioritization of action, based on regulation by evidence and risk management, with focus and orientation by results; acting in a responsive manner, with the adoption of regimes proportional to the identified risk and the position of the regulated agent; and provision of transparency, feedback and self-regulation mechanisms (Anatel 2021b).

It is evident that the success of the Anatel (and that of any regulatory authority with scarce resources) depends on properly prioritizing its actions, as made clear in Article 5 of RER. The prioritization methodology is described in Article 10 of RER. According to Article 10, prioritization must observe, among other aspects, the direct impact on consumers and regional characteristics, as well as particularities in the provision of services (Anatel 2021b).

We recall that the quality of service for the regulator depends on the disparity between the regulator perception about the service and the regulator expectation about the service. The former can be determined by the consumer complaints rate by municipality and operator. The latter will then be given by the regulator expectation about the consumer complaints rate by municipality and operator.

Consider a graph $\mathcal{G}$ in which each vertex $v_i$ represents one of the 5570 municipalities in Brazil, and the signal $x_i$ at vertex $v_i$ represents the number of consumer complaints per 100,000 consumers at municipality $v_i$ for one of the operators.

Anatel listens to consumers through four channels: in person, by letter, by phone, and through a mobile phone application. Only in 2020, Anatel received more than 2.96 million complaints (Anatel 2021c). Anatel also knows the number of consumers for each pair municipality-operator. Hence, the information $x$ is available.

The adjacency matrix $W$ was defined to adequately model the regulator expectations about the consumer complaint rate at each municipality. To build the adjacency matrix $W$, we used the available information about the municipalities $v_j$ that serve as reference for each municipality $v_i$. We then connected each vertex $v_j$ in this group to $v_i$, so that municipalities $v_j$ became predecessors of $v_i$. The weights $w_{ij}$ define the influences of each $v_j$ on $v_i$.

Commonly, one expects that services be initially provided to large urban centers, and then be progressively made available to smaller centers (Rauhut & Humer 2020). In Brazil, this becomes clear when one observes Anatel's auctions of radio-frequency for implementation and evolution of the PCS (Anatel 2007; 2010; 2012; 2021a). These auctions classified Brazilian municipalities according to their total population and established commitments to implement services initially in the larger centers and later on in smaller centers. The commitments for service evolution were also established so that the service progressively develops from the larger urban centers to the smaller centers. Larger urban centers, therefore, serve as references for smaller urban centers when it comes to the provision of telecommunications services.

In 2020, the Brazilian Institute of Geography and Statistics (IBGE) published the study Regions of Influence of Cities 2018, in which it identifies and analyzes the Brazilian urban network, establishing the hierarchy and regions of influence of urban centers. In the IBGE study, urban centers can be composed of not only one but several municipalities that are inseparable as urban units. To establish the hierarchy and regions of influence of urban centers, the IBGE considered the long-distance relationships between these centers, generated by command and control relations, and the urban relations of proximity, generated by the search for goods and services (IBGE 2020b). The Brazilian urban network is characterized by small urban centers being influenced by one, or more, larger urban centers. At the top of the network are the 15 Brazilian metropolises, which are influenced by one or more metropolises (IBGE 2020b).

To determine the weights of the influences between urban centers, we used results provided by IBGE in which relations between centers were classified as being of 1st, 2nd, or 3rd order (IBGE 2020a). For influences in which an urban center looks for goods or services in another urban center, the influence weight was calculated as an average of the existing relations between the two centers in ten areas researched by the IBGE, namely: clothing and footwear, furniture and electronics, low- and medium-complexity healthcare, high-complexity healthcare, higher education, cultural activities, sports activities, airport, newspapers and public transportation. 2nd and 3rd order relations, related to the search for goods and services, were weighted as corresponding to 95% and 90% of the importance of the 1st order relations, respectively, as indicated by IBGE in its study (IBGE 2020b). For influences between metropolises that do not seek goods and services in other urban centers but are linked to other metropolises, the weights of the influences were calculated as an average of the relations between metropolises in one of the four themes researched by IBGE, namely: public management, business management, road and waterway links and airway links. For the influences between metropolises, the 2nd and 3rd order relations were weighted as corresponding to 50% and 33.33%, respectively, of the importance of the 1st order relations, as indicated by IBGE in its study (IBGE 2020b). The IBGE study also identified urban centers that were not metropolises, but nevertheless did not seek goods and services in other centers. In these cases, it was considered that all the influence came from the metropolis to which each one of them was linked by the IBGE.

As an urban center in the IBGE study could be formed by several municipalities, urban centers were classified as influencing or influenced urban centers. Then, the influence of an urban center on a municipality was distributed among the municipalities in the influencing urban center proportionally to their populations.

Figure 1 illustrates the connectivity of São Paulo with the other municipalities in Brazil, totaling 912 edges. The actual graph representing all connections between Brazilian municipalities, as described above, is much more complex, with 5,570 vertices and 138,382 edges.

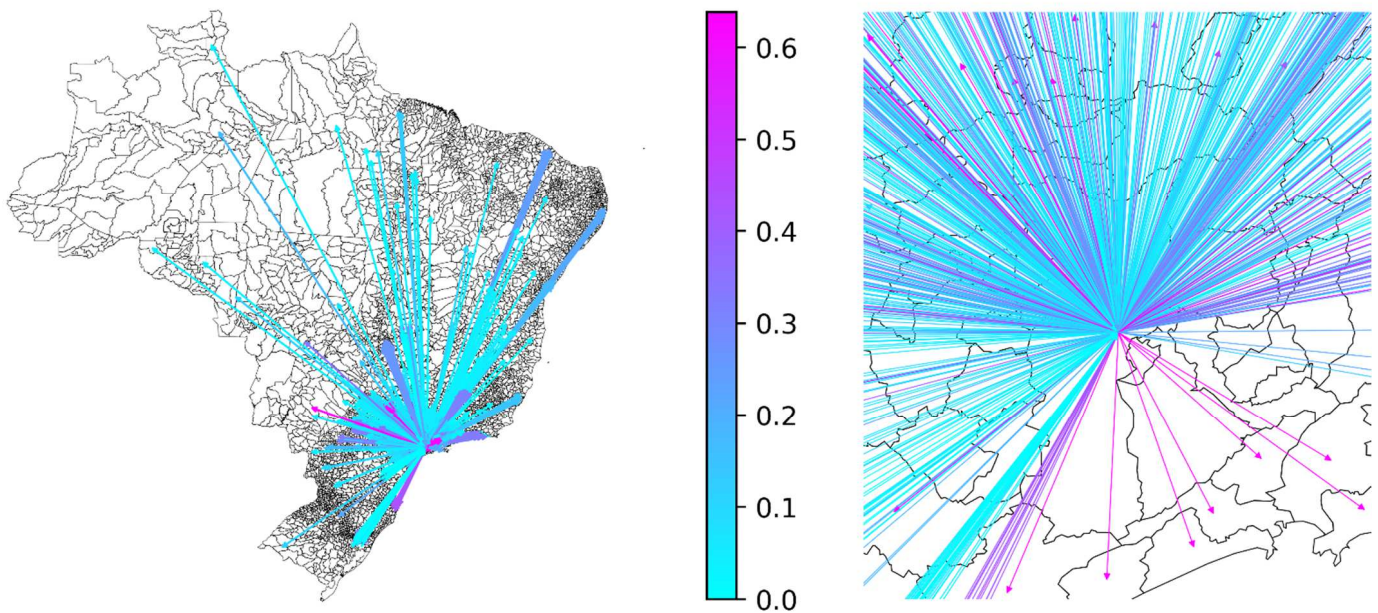

**Figure 1.** Influences between São Paulo and the others municipalities in Brazil and focus on São Paulo.

The complete graph, characterized by the set $\mathcal{V}$ of vertices and by adjacency matrix $\boldsymbol{W}$ was then used to determine the quality of PCS from the regulator's point of view at each municipality $v_i$. The PCS quality is represented by the discrepancy $d_i$ at each vertex. To avoid noisy transients characteristic of daily rates of complaints, we have considered a 28-day moving average of the consumer complaint rates as graph signal (Anatel 2022, unpublished data). The vertex discrepancy vector $\boldsymbol{d}$ was determined for each of the four major PCS operators in Brazil, named here from A to D, with $\boldsymbol{x}$ being the vector formed by the 28-day moving averages of complaint rate at each municipality. The number of consumers by municipality and operator were updated monthly. In this way, the graph signal $\boldsymbol{x}$ was formed by the values $x_i$ given by the 28-day moving average of complaints per 100,000 consumers at each municipality $v_i$.

Suppose that Anatel wants to carry out ten inspections per month, being five in municipalities with total population greater than 500,000 inhabitants and five in municipalities with total population between 200,000 and 500,000 inhabitants. The focus is on municipalities-operators whose PCS qualities were the worst in the segmentation by municipality and operator at the end of each month.

Table 1 shows data for the five worst municipality-operator pairs at the end of each month of 2021 for the first group. Letters A-D besides the city name identify the corresponding operator. Table 2 shows the classification for the second group. The numbers separated by dashes provide two different quality rankings of municipality-operator pairs. Classifications are from worse (smaller rank numbers) to better (higher rank numbers). For comparison, we list the classifications according to two criteria: i) based on the vertex divergence $d_i$ (number before dash) and ii) based on the average complaints per 100,000 consumers (numbers after dash). For instance, a municipality-operator pair with ranks 1-15 means that it was ranked the worst (rank 1) for that month according to the discrepancy criterion, and ranked the 15th worst according to the average number of complaints. We shade the background for those municipalities-operators among the five worst by the vertex divergence criterion when, by the average complaint rate criterion, the classification is not among the five worst. This is done to facilitate comparison. For instance, the first row of Table 1 shows that Belford-Roxo - D had the worst vertex divergence in January 2021, indicating high priority for inspection by the proposed method. However, it would not be selected for inspection if just the average complaint rate at this municipality-operator was to be considered, since it would have been classified in the 15th position and just five inspections would be carried out. Similar situations occur for one third of the cases shown in Table 1 and for about one half of the cases shown in Table 2. In both tables, the signaled differences happen spread throughout the whole year. We also note that some differences in classifications are very significant, evidencing a considerable difference in inspection prioritization strategies depending on the chosen methodology. This is because the isolated consumer complaint rates do not take into consideration the performance at a municipality relatively to the performance at other municipalities. Analysis of these two tables clearly shows the importance of considering performances in the context of municipality influences, which is possible using the proposed GSP framework and the vertex divergence metric.

**Table 1.** Inspection order for operators and municipalities with more than 500,000 inhabitants based on vertex discrepancy (before dash) and on the average consumer complaint rate (after dash). The smaller the number, the greater the priority for inspection. The gray background indicates that the priority for inspection would not be among the top 5 if the consumer complaint rate were the criterion for targeting the inspections.

| Vertex | Jan | Feb | Mar | Apr | May | Jun | Jul | Aug | Sep | Oct | Nov | Dec |
|---|---|---|---|---|---|---|---|---|---|---|---|---|
| Belford Roxo - D | 1-15 | | | | | | | 4-1 | | 3-3 | | |
| Contagem - A | 2-1 | | 2-1 | 4-1 | 5-1 | 2-1 | 4-1 | | | 2-1 | | |
| Belo Horizonte - C | 3-2 | 2-2 | 3-5 | | 4-3 | 5-3 | 5-5 | | | | | 4-15 |
| Belford Roxo - B | 4-17 | | | 3-14 | | | 3-3 | | | | | 1-5 |
| Contagem - C | 5-3 | 4-3 | | | 3-2 | | 2-2 | 5-10 | | 5-10 | 3-11 | |
| Guarulhos - A | | 1-4 | 1-2 | 2-3 | 1-4 | 1-2 | 1-4 | 2-2 | 2-6 | 4-4 | 5-5 | 3-3 |
| Santo André - A | | 3-5 | 4-6 | | 2-5 | | | | | | 2-4 | |
| Jaboatão dos Guararapes - D | | 5-15 | | | | | | | | | | |
| São Bernardo do Campo - A | | | 5-7 | 5-6 | | | | | | | | |
| Aparecida de Goiânia - B | | | | 1-2 | | 3-4 | | 3-4 | 1-1 | | 1-2 | 2-7 |
| Contagem - D | | | | | | 4-14 | | | | | | |
| Caxias do Sul - C | | | | | | | | 1-6 | 5-19 | | | 5-26 |
| Osasco - A | | | | | | | | | 3-12 | | | |
| Campos dos Goytacazes - A | | | | | | | | | 4-3 | | | |
| Aracaju - B | | | | | | | | | | 1-26 | | |
| Curitiba - B | | | | | | | | | | 4-13 | | |

**Table 2.** Inspection order for operators and municipalities with more than 200,000 and less than 500,00 inhabitants based on vertex discrepancy (before dash) and on the average consumer complaint rate (after dash). The smaller the number, the greater the priority for inspection. The gray background indicates that the priority for inspection would not be among the top 5 if the consumer complaint rate were the criterion for targeting the inspections.

| Vertex | Jan | Feb | Mar | Apr | May | Jun | Jul | Aug | Sep | Oct | Nov | Dec |
|---|---|---|---|---|---|---|---|---|---|---|---|---|
| *Águas Lindas de Goiás - D* | 1-1 | 1-1 | 1-1 | 1-1 | 1-1 | 2-1 | 1-1 | 1-1 | 1-1 | 1-1 | 1-1 | 1-1 |
| *Sobral - B* | 2-11 | 2-9 | | 2-2 | 2-15 | 1-3 | | 2-2 | 2-2 | | | |
| *Arapiraca - B* | 3-71 | | | 3-4 | 5-60 | 5-81 | | | | | 5-56 | 3-18 |
| *Itabuna - A* | 4-49 | 3-31 | 3-32 | | 4-2 | 3-39 | 2-57 | 3-56 | 3-15 | | | 2-16 |
| *Franca - D* | 5-27 | 4-10 | 4-22 | | 3-3 | 4-5 | 4-4 | | | 3-3 | 4-5 | 4-5 |
| *Mossoró - D* | | 5-51 | | | | | | | | | | |
| *Campina Grande - D* | | | 2-104 | 4-101 | | | | | | 4-25 | 2-8 | |
| *Arapiraca - A* | | | 5-24 | 5-5 | | | | 4-6 | 4-4 | 2-4 | 3-3 | 5-14 |
| *Criciúma - B* | | | | | | | 3-3 | | | | | |
| *Sobral - D* | | | | | | | 5-2 | | | | | |
| *Mogi das Cruzes - A* | | | | | | | | 5-4 | | | | |
| *Rio Branco - C* | | | | | | | | | 5-234 | | | |
| *Mauá - A* | | | | | | | | | | 5-5 | | |

Another important application of the proposed GSP-based analysis framework is the detection of changes in the quality of the PCS. For this, it is enough that the sequence of values $d_i$ for each municipality-operator pair be processed by some change detection algorithm, such as CUSUM (Basseville & Nikiforov 1993). As an example, the blue curve in Figure 2 shows the evolution of the discrepancy of operator B in the municipality of Criciúma/SC. The red curve shows the result of applying the CUSUM algorithm to this vertex discrepancy. It clear that CUSUM will evidence large changes in the vertex discrepancy, which can serve as an alarm for the regulator to determine the need for an inspection of that municipality-operator pair.

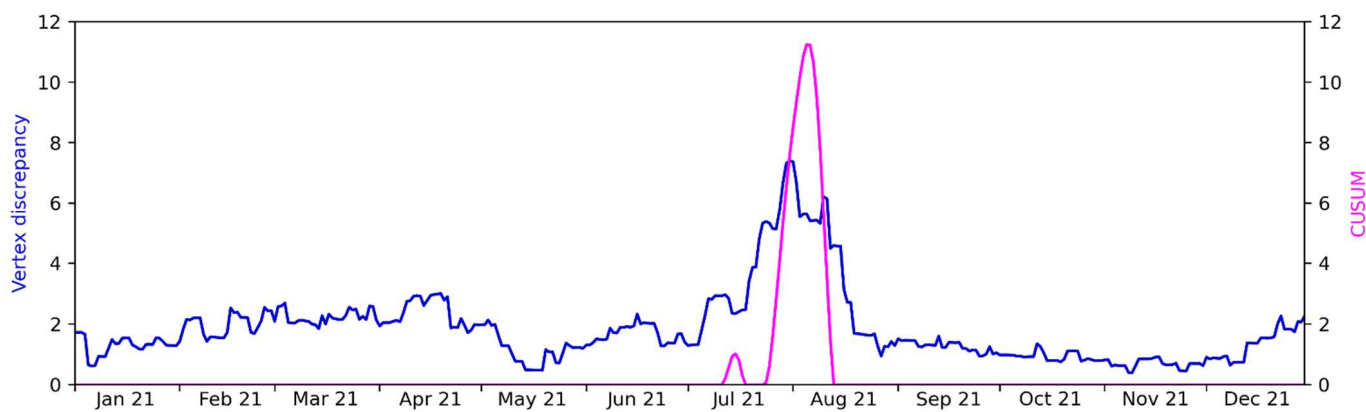

**Figure 2.** Vertex discrepancy and CUSUM of operator B at Criciúma/SC.

## 6. Discussion

Despite considerable advances, risk-based regulation has not been widely implemented in the world, even where regulation is said to be risk-based (OECD 2021b). The lack of risk analysis is especially harmful for monitoring the market and targeting inspections, and without the latter no type of regulation will yield good results (Pecaric 2017), even for the most important and common subject for the majority of regulators. To overcome this difficulty, it is necessary to take advantage of easier and cheaper access to data and data processing (OECD 2021a) to ensure that the data are reliable and the processing methods are objective, impartial, transparent, explainable, simple to implement and with low computational cost.

If the objective is the wide use of data processing methods by regulators to perform risk analysis, it is necessary to be aware that inherent characteristics of machine learning methods conflict with essential characteristics of processing methods that could be used to achieve that objective. Machine learning methods always train their models from the available data and then make analysis and predictions (Theodoridis 2015). Moreover, machine learning methods always deal with uncertainty, either by the limitation in the data available for training the model, by the stochasticity of the model or by the incapacity to model (Goodfellow *et al*. 2016; Murphy 2012). For risk analysis of wide application in regulation, the dataset, although big, is always limited and the systems to be modeled are always complex. Therefore, machine learning methods are subject to errors related to the choice of the model structure and the training step of the chosen model. The complexity of the model to be used in the machine learning method, added to the imprecision of the model due to limited training, results in an undetermined solution, which is difficult to implement and to explain, and without guarantees of impartiality. In this case, opting for machine learning solutions may mean to use solutions that are more difficult to develop, implement, explain, communicate and, after all this effort, are still fragile and failure-prone (Goodfellow *et al*. 2016). Although we are experiencing the 3rd hype of artificial intelligence, we have already experienced two winters on the theme, in the 70s and 90s, and parsimony is necessary in the use of these methods (Chollet 2021), especially regarding the wide use of machine learning methods in regulation (Telecommunications Policy 2020; Regulation & Governance 2022).

The proposed solution has the characteristics necessary for data processing methods to be widely used by regulators in risk analysis. It is objective, impartial, transparent, explainable and simple to implement. In the proposed solution, the resulting ranking occurs in relation to the chosen model representing the influence network among the vertices of the graph, that is, it is only necessary to deal with this modeling error. The proposed solution also demonstrates that errors may be being made in targeting inspections when a consumer complaint rates ranking is used directly for this purpose. Furthermore, working with the vertex discrepancies has low computational complexity, avoiding problems with singular or poorly conditioned matrices or with the need for high numerical precision for the calculations, allowing to work with very large graphs. Hence, formulating the problem using the proposed method leads to an effective solution.

It is important to note that the development of the vertex discrepancy did not consider what vertices $v_i$ of the graph actually represent. The weights $w_{ij}$ of the edges between the vertices and the signal $x_i$ at each vertex on the graph are also defined as general entities, without a necessary association to the particular problem about consumer complaints. The vertex discrepancy can, therefore, be used for any analysis where the interest is to calculate how proportionally discrepant the signal observations at graph vertices are from their predicted values, if the graph signal transformation by the adjacency matrix properly models the values that are expected for the vertices. Hence, the proposed GSP-based framework can be applied to different data segmentations and objectives. In certain applications, one may be interested in studying the performance of the graph signal by group of vertices. Extending the use of vertex discrepancy to group discrepancy is straightforward.

## 7. Conclusion

Aware of the fact that the risk-based regulation has not been widely implemented in the world, even where regulation is said to be risk-based, in this article we drew attention to the need and proposed a data-driven processing method which is objective, impartial, transparent, explainable and simple to implement, with low computational cost, to solve the problems of market monitoring and inspection targeting. We showed that consumer policy is a topic common to the majority of regulators, and proposed a graph signal processing-based method for targeting inspections. Consumer complaint data of the regulatory authority for the Brazilian telecommunications market were used to illustrate the implementation of the proposed method, evidencing that errors may be being made in targeting the inspections when a consumer complaint rates ranking is used directly for this purpose. To the best of our knowledge, this is the first methodology based on graph signal processing proposed for targeting regulatory inspections. We show why inherent characteristics of machine learning methods conflict with essential characteristics of data processing methods that could be widely used by regulators in risk analysis. With the discussion made in this article, we hope that new methods for risk analysis, especially to solve the problems of monitoring the market and targeting inspections in a data-driven way, can be proposed, widely debated and possibly adopted by regulators around the world.

**Acknowledgments**

Celso H. H. Ribas is a regulatory specialist at the National Telecommunications Agency, the independent regulatory authority for the Brazilian telecommunications market. This is an academic work without any competing interest. The views expressed here are the author's and do not reflect those of the institutions to which the authors belong.